\documentclass[journal]{IEEEtran}
\usepackage{times}
\usepackage{epsfig}
\usepackage{graphicx}
\usepackage{amsmath}
\usepackage{amssymb}
\usepackage{bm}
\usepackage{booktabs}
\usepackage{xcolor}

\usepackage[breaklinks=true,bookmarks=false]{hyperref}

\usepackage{etoolbox}
\makeatletter
\patchcmd{\@makecaption}
  {\scshape}
  {}
  {}
  {}
\makeatother

\newcommand{\etal}{\mbox{\emph{et al.\ }}}

\begin{document}

\title{Statistics of Deep Generated Images}

\author{Yu~Zeng, Huchuan~Lu, and Ali~Borji
\thanks{Y. Zeng is with School of Information and Communication Engineering at Dalian University of Technology, Dalian, China. E-mail: zengyu@mail.dlut.edu.cn.}
\thanks{H. Lu is with School of Information and Communication Engineering at Dalian University of Technology, Dalian, China. E-mail: lhchuan@dlut.edu.cn.}
\thanks{A. Borji is with the Center for Research in Computer Vision and Computer Science Department at the University of Central Florida, Orlando, FL. aborji@crcv.ucf.edu}

\thanks{Manuscript received xx 2017.}

}
\markboth{Submitted to IEEE Transactions on Image Processing,~Vol.~xx, No.~xx, xx}%
{Statistics of Deep Generated Images}
\maketitle
\begin{abstract}
Here, we explore the low-level statistics of images generated by state-of-the-art deep generative models. 
First, Variational auto-encoder (VAE~\cite{kingma2013auto}), Wasserstein generative adversarial network (WGAN~\cite{arjovsky2017wasserstein}) and deep convolutional generative adversarial network (DCGAN~\cite{radford2015unsupervised}) are trained on the ImageNet dataset and a large set of cartoon frames from animations. Then, for images generated by these models as well as natural scenes and cartoons, statistics including mean power spectrum, the number of connected components in a given image area, distribution of random filter responses, and contrast distribution are computed. Our analyses on training images support current findings on scale invariance, non-Gaussianity, and Weibull contrast distribution of natural scenes. We find that although similar results hold over cartoon images, there is still a significant difference between statistics of natural scenes and images generated by VAE, DCGAN and WGAN models. In particular, generated images do not have scale invariant mean power spectrum magnitude. Inspecting how well the statistics of deep generated images match the known statistical properties of natural images, such as scale invariance, non-Gaussianity, and Weibull contrast distribution, can a) reveal the degree to which deep learning models capture the essence of the natural scenes, b) provide a new dimension to evaluate models, and c) allow possible improvement of image generative models (e.g., via defining new loss functions). 
\end{abstract}
\begin{IEEEkeywords}
generative models, image statistics, convolutional neural networks, deep learning.
\end{IEEEkeywords}
\IEEEpeerreviewmaketitle


\section{Introduction and Motivation}
\IEEEPARstart{G}ENERATIVE models are statistical models that attempt to explain observed data by some underlying hidden (i.e., latent) causes~\cite{hyvarinen2009natural}. Building good generative models for images is very appealing for many computer vision and image processing tasks. Although a lot of previous effort has been spent on this problem and has resulted in many models, generating images that match the qualities of natural scenes remains to be a daunting task.

\begin{figure}[t]
\begin{center}
   \includegraphics[width=\linewidth]{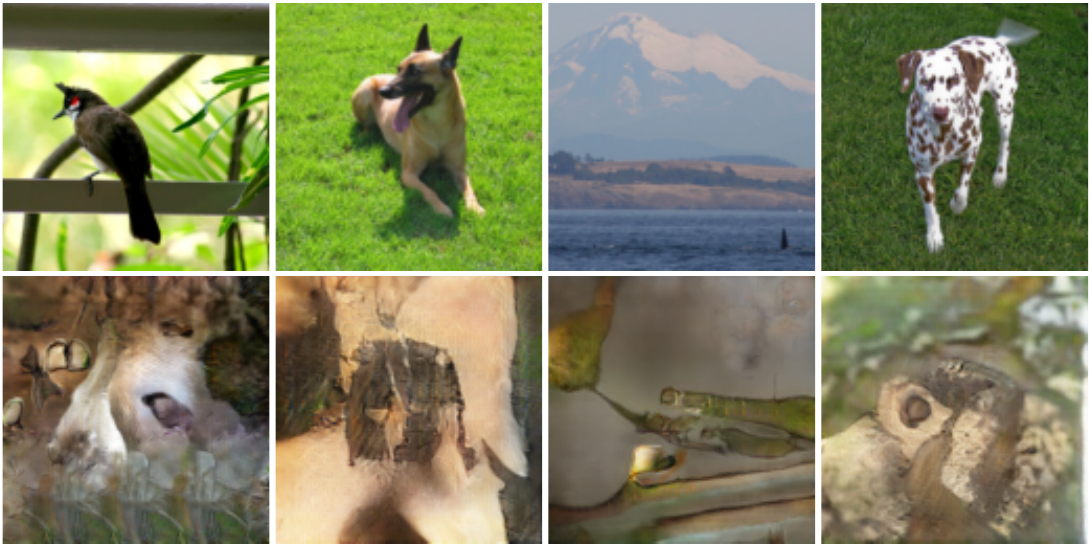}
\end{center}
   \vspace{-10pt}
   \caption{Top: training images from ImageNet dataset~\cite{deng2009imagenet}. Bottom: images generated by DCGAN~\cite{radford2015unsupervised}.}
   \label{samples}
   \vspace{-10pt}
\end{figure}

There are two major schemes for the design of image generative models. The first one is based on the known regularities of natural images and aims at satisfying the observed statistics of natural images. Examples include the Gaussian MRF model~\cite{mumford2001stochastic}, for the $1/f$-power law, and the Dead leaves model~\cite{zhu2003statistical} for the scale invariant property of natural images. These models are able to reproduce the empirical statistics of natural images well~\cite{lee2001occlusion}, but images generated by them do not seem very realistic. The second scheme is data-driven. It assumes a flexible model governed by several parameters, and then learns the parameters from training data. Thanks to large image datasets and powerful deep learning architectures, the second scheme has been adopted in most of the recent image generation models. Typical examples include variational autoencoders (VAE)\cite{kingma2013auto} and generative adversarial networks (GAN)\cite{goodfellow2014generative}. Utilizing convolutional neural networks~\cite{krizhevsky2012imagenet}, as building blocks, and training on tens of thousands of images, deep generative models are able to generate plausible images, as shown in the second  row of Figure~\ref{samples}. 

On the one hand, despite the promise of deep generative models to recover the true distribution of images, formulating these models usually involves some sort of approximation. For instance, the variational auto-encoder (VAE)\cite{kingma2013auto} aims at estimating an explicit probability distribution through maximum likelihood, but the likelihood function is intractable. So a tractable lower bound on log-likelihood of the distribution is defined and maximized. The generative adversarial network (GAN)\cite{goodfellow2014generative} can recover the training data distribution when optimized in the space of arbitrary functions, but in practice, it is always optimized in the space of the model parameters. Therefore, there is basically no theoretical guarantee that the distribution of images by generative models is identical to that of natural images. On the other hand, images generated by deep generative models, hereinafter referred to as deep generated images, indeed seem different from natural images such that it is easy for humans to distinguish them from natural images~\cite{denton2015deep}. Please see the first and the second rows of Figure~\ref{samples}. It remains unclear whether deep generative models can reproduce the empirical statistics of natural images. 

Driven by this motivation, we take the generative adversarial networks and variational auto-encoders as examples to explore statistics of deep generated images with respect to natural images in terms of scale invariance, non-Gaussianity, and Weibull contrast distribution. These comparisons can reveal the degree to which the deep generative models capture the essence of the natural scenes and  guide the community to build more efficient generative models. In addition, the current way of assessing image generative models are often based on visual fidelity of generated samples using human inspections~\cite{theis2015note}. As far as we know, there is still not a clear way to evaluate image generative models~\cite{im2016generating}. We believe that our work will provide a new dimension to evaluate image generative models.

Specifically, we first train a Wasserstein generative adversarial network (WGAN~\cite{arjovsky2017wasserstein}), a deep convolutional generative adversarial network (DCGAN~\cite{radford2015unsupervised}), and a variational auto-encoder (VAE~\cite{kingma2013auto}) on the ImageNet dataset. The reason for choosing ImageNet dataset is that it contains a large number of photos from different object categories. We also collect the same amount of cartoon images to compute their statistics and to train the models on them, in order to: 1) compare statistics of natural images and cartoons, 2) compare statistics of generated images and cartoons, and 3) check whether the generative models work better on cartoons, since cartoons have less  texture than natural images. As far as we know, we are the first to investigate statistics of cartoons and deep generated images. Statistics including luminance distribution, contrast distribution, mean power spectrum, the number of connected component with a given area, and distribution of random filter responses will be computed. 

Our analyses on training natural images confirm existing findings of scale invariance, non-Gaussianity, and Weibull contrast distribution on natural image statistics. We also find non-Gaussianity and Weibull contrast distribution in VAE, DCGAN and WGAN's generated natural images. However, unlike real natural images, neither of the generated images have scale invariant mean power spectrum magnitude. Instead, the deep generative models seem to prefer certain frequency points, on which the power magnitude is significantly larger than their neighborhood.

\section{Related Work}
In this section, we briefly describe recent work that is closely related to this paper, including important findings in the area of natural image statistics and recent developments on deep image generative models. 

\subsection{Natural Image Statistics}
Research on natural image statistics has been growing rapidly since the mid-1990’s\cite{hyvarinen2009natural}. The earliest studies showed that the statistics of the natural images remains the same when the images are scaled (i.e., scale invariance)\cite{srivastava2003advances, zhu2003statistical}. For instance, it is observed that the average power spectrum magnitude $A$ over natural images has the form of $A(f) = 1/f^{-\alpha},~\alpha \approx 2$ (See for example~\cite{deriugin1956power, cohen1975image, burton1987color, field1987relations}). It can be derived using the scaling theorem of the Fourier transformation that the power spectrum magnitude will stay the same if natural images are scaled by a factor~\cite{zoran2013natural}. Several other natural image statistics have also been found to be scale invariant, such as the histogram of log contrasts~\cite{ruderman1994statistics2}, the number of gray levels in small patches of images~\cite{geman1999invariant}, the number of connected components in natural images~\cite{alvarez1999size}, histograms of filter responses, full co-occurrence statistics of two pixels, as well as joint statistics of Haar wavelet coefficients. 

Another important property of natural image statistics is the non-Gaussianity~\cite{srivastava2003advances, zhu2003statistical, wainwright1999scale}. This means that marginal distribution of almost any zero mean linear filter response on virtually any dataset of images is sharply peaked at zero, with heavy tails and high kurtosis (greater than 3 of Gaussian distributions)~\cite{lee2001occlusion}. 

In addition to the two well-known properties of natural image statistics mentioned above, recent studies have shown that the contrast statistics of the majority of natural images follows a Weibull distribution~\cite{ghebreab2009biologically}. Although less explored, compared to the scale invariance and non-Gaussianity of natural image statistics, validity of Weibull contrast distribution has been confirmed in several studies. For instance, Geusebroek~\etal~\cite{geusebroek2005six} show that the variance and kurtosis of the contrast distribution of the majority of natural images can be adequately captured by a two-parameter Weibull distribution. It is shown in~\cite{scholte2009brain} that the two parameters of the Weibull contrast distribution cover the space of all possible natural scenes in a perceptually meaningful manner. Weibull contrast distribution also has been applied to a wide range of computer vision and image processing tasks. Ghebreab~\etal\cite{ghebreab2009biologically} propose a biologically plausible model based on Weibull contrast distribution for rapid natural image identification, and Yanulevskaya~\etal\cite{yanulevskaya2011image} exploit this property to predict eye fixation location in images, to name a few. 

\subsection{Deep Generative Models}
Several deep image generative models have been proposed in a relatively short period of time since 2013. As of this writing, variational autoencoders (VAE) and generative adversarial networks (GAN) constitute two popular categories of these models. VAE aims at estimating an explicit probability distribution through maximum likelihood, but the likelihood function is intractable. So a tractable lower bound on log-likelihood of the distribution is defined and maximized. For many families of functions, defining such a bound is possible even though the actual log-likelihood is intractable. In contrast, GANs implicitly estimate a probability distribution by only providing samples from it. Training GANs can be described as a game between a generative model $G$ trying to estimate data distribution and a discriminative model $D$ trying to distinguish between the examples generated by $G$ and the ones coming from actual data distribution. In each iteration of training, the generative model learns to produce better fake samples while the discriminative model will improve its ability of distinguishing real samples.

It is shown that a unique solution for $G$ and $D$ exists in the space of arbitrary functions, with $G$ recovering the training data distribution and $D$ equal to $\frac{1}{2}$ everywhere~\cite{goodfellow2014generative}. In practice, $G$ and $D$ are usually defined by multi-layer perceptrons (MLPs) or convolutional neural networks (CNNs), and can be trained with backpropagation through gradient-based optimization methods. However, in this case, the optimum is approximated in the parameter space instead of the space of arbitrary functions. Correspondingly, there is no theoretical guarantee that the model's distribution is identical to the data generating process~\cite{goodfellow2014generative}. 

Generally speaking, image samples generated by GANs and VAEs look quite similar to the real ones, but there are indeed some differences. Figure~\ref{samples} shows samples of training images from ImageNet, and images generated by a popular implementation of GANs, termed as DCGAN~\cite{radford2015unsupervised}. As humans, we can easily distinguish fake images from the real ones. However, it is not so easy to tell how different deep generated images are from the real ones, and whether deep generative models, trained on a large number of images, capture the essence of the natural scenes. We believe that answering how well the statistics of the deep generated images match with the known statistical properties of natural images, reveals the degree to which deep generative models capture the essence of the natural scenes. Insights can be gained from this work regarding possible improvements of image generative models.

\section{Data and Definitions}
In this section, we introduce data, definitions, and symbols that will be used throughout the paper.

\subsection{Natural Images, Cartoons and Generated Images}
\begin{figure}[t]
\begin{center}
\includegraphics[width=\linewidth]{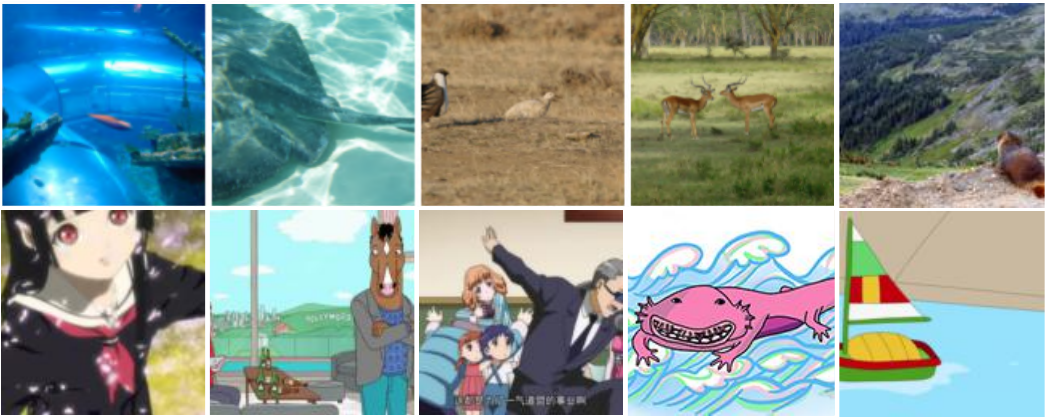}
\end{center}
   \vspace{-10pt}
   \caption{Examples of natural images from the ImageNet dataset~\cite{deng2009imagenet} (top row), and our collected cartoon images (bottom row).}
   \label{traing_images}
\end{figure}

We choose 517,400 out of 1,300,000 pictures of ImageNet~\cite{deng2009imagenet} dataset as our natural image training set. These images cover 398 classes of objects, and each class contains 1,300 images. The cartoon training images include 511,460 frames extracted from 303 videos of 73 cartoon movies (i.e., multiple videos per movie). These two sets are used to train the deep generative models to generate natural images and cartoons. All training images are cropped around the image center. Each image has $128 \times 128$ pixels. Figure~\ref{traing_images} shows some examples of the natural and cartoon training images. 

\begin{figure}[t]
\begin{center}
\includegraphics[width=\linewidth]{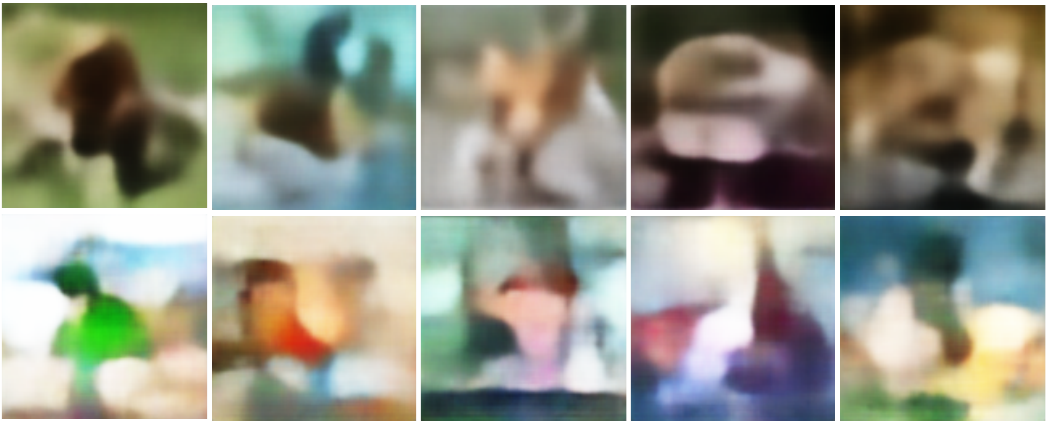}
\end{center}
   \vspace{-10pt}
   \caption{Examples of natural (top) and cartoon images (bottom) generated by the VAE model~\cite{kingma2013auto}.}
   \label{vae_images}
\end{figure}

\begin{figure}[t]
\begin{center}
\includegraphics[width=\linewidth]{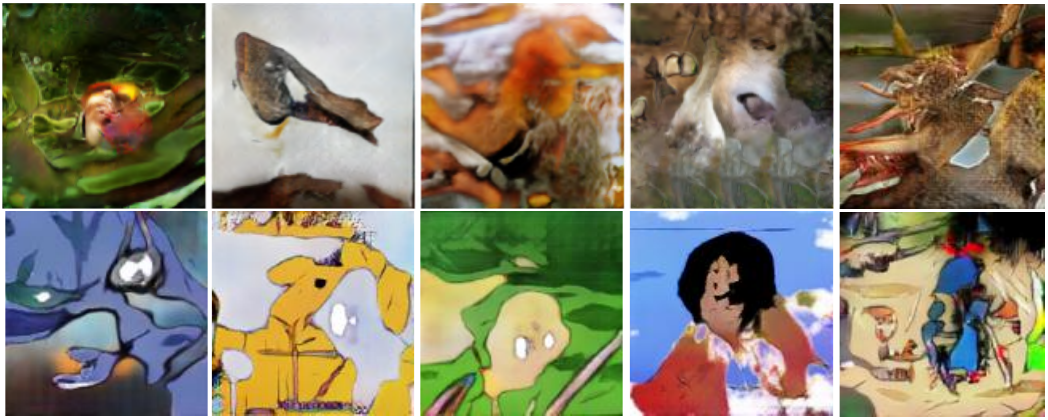}
\end{center}
   \vspace{-10pt}
   \caption{Examples of natural (top) and cartoon images (bottom) generated by the DCGAN model~\cite{radford2015unsupervised}.}
   \label{dcgan_images}
\end{figure}

\begin{figure}[t]
\begin{center}
\includegraphics[width=\linewidth]{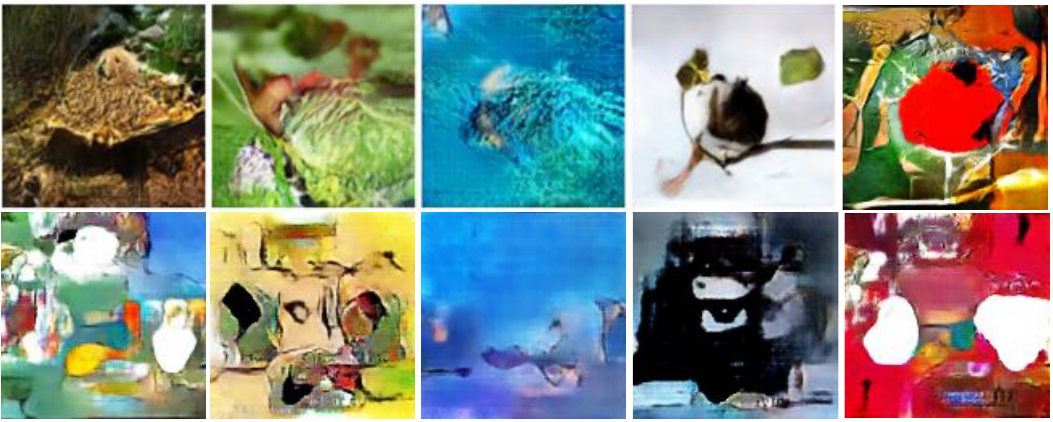}
\end{center}
   \vspace{-10pt}
   \caption{Examples of natural (top) and cartoon images (bottom) generated by the WGAN model~\cite{arjovsky2017wasserstein}.}
   \label{wgan_images}
\end{figure}

Several variants of deep generative models have been proposed. Since  it is nearly impossible to consider all models, here we focus on three leading models including VAE,  DCGAN and WGAN for our analysis. DCGAN refers to a certain type of generative adversarial networks with the architecture proposed by Radford~\etal~\cite{radford2015unsupervised} and the cost function proposed by Goodfellow~\etal~\cite{goodfellow2014generative}. WGAN refers to the model with the architecture proposed by Radford~\etal~\cite{radford2015unsupervised} and the cost function proposed by Arjovsky~\etal~\cite{arjovsky2017wasserstein}. VAE approach, proposed by Kingma~\etal~\cite{kingma2013auto}, consists of fully connected layers which are not efficient in generating large images. Therefore, we replace the architecture of the original VAE with the convolutional architecture proposed by Radford~\etal~\cite{radford2015unsupervised}. In short, the DCGAN, WGAN and VAE models used in this paper have the same architecture. Their difference lies in their loss functions. The generated images considered in this work have the size of $128 \times 128$ pixels. Examples of images generated by VAE, DCGAN and WGAN are shown in Figures~\ref{vae_images}, ~\ref{dcgan_images} and~\ref{wgan_images}, respectively.

\subsection{Kurtosis and Skewness}
Kurtosis is a measure of the heaviness of the tail of a probability distribution. A large kurtosis indicates that the distribution has a sharp peak and a heavy tail. Skewness measures asymmetry of a probability distribution with respect to the mean. A positive skewness indicates the mass of the distribution is concentrated on the values less than the mean, while a negative skewness indicates the opposite. The kurtosis and skewness of a random variable $X$ are defined as:
\begin{equation}
\mathcal{K} = \frac{E[(X-\mu)^4]}{\sigma^4}, 
\end{equation}
\begin{equation}
\mathcal{S} = \frac{E[(X-\mu)^3]}{\sigma^3}, 
\end{equation}
where $\mu$ is the mean, $\sigma$ is the standard deviation, and $E[\cdot]$ denotes the mathematical expectation. 

\subsection{Luminance}
Since training and deep generated images are RGB color images, first we convert them to grayscale using the formula as for CCIR Rec. 601, a standard for digital video, as follow:
\begin{equation}
Y = 0.299 R + 0.587 G + 0.114 B.
\end{equation}
It is a weighted average of R, G, and B to tally with human perception. Green is weighted most heavily since human are more sensitive to green than other colors~\cite{kanan2012color}. The grayscale value $Y(i, j)$ of the pixel at position $(i, j)$ is taken as its luminance. Following~\cite{geisler2008visual}, in this work, we deal with the normalized luminance $I$ within a given image which is defined by dividing the luminance $Y(i, j)$ at each pixel by the average luminance over the whole image:
\begin{equation}
I(i, j) = \frac{Y(i, j)}{\frac{1}{HW}\sum_{i, j} Y(i, j)}, 
\end{equation}
where $H$ and $W$ are the height and width of the image, respectively. Averaging the luminance histograms across images gives the distribution of luminance. 

As a fundamental feature encoded by biological visual systems, luminance distribution within natural images has been studied in many works. It has been observed that this distribution is approximately symmetric on a logarithmic axis and hence positively skewed on a linear scale~\cite{geisler2008visual}. In other words, relative to the mean luminance, there are many more dark pixels than light pixels. One reason is the presence of the sky in many images, which always has high luminance, causing the mean luminance to be greater than the luminance of the majority of pixels. 

\subsection{Contrast Distribution}
\label{sec_contrast}
Distribution of local contrast within images has been measured using various definitions of contrast. In this work, we use the gradient magnitude calculated by Gaussian derivative filters to define local contrast of an image, as in~\cite{scholte2009brain,ghebreab2009biologically,yanulevskaya2011image}. These contrast values have been shown to follow a Weibull distribution~\cite{ghebreab2009biologically}:
\begin{equation}
p(x) = \frac{\gamma}{\beta} (\frac{x}{\beta})^{\gamma-1} e^{-(\frac{x}{\beta})^\gamma}.
\label{wdis}
\end{equation} 

Images are firstly converted to a color space that is optimized to match the human visual system color representation~\cite{yanulevskaya2011image}:
\begin{equation}
\begin{bmatrix}
   E_1 \\
   E_2 \\
   E_3 
  \end{bmatrix}
  =
\begin{bmatrix}
  0.06& 0.63& 0.27 \\
   0.3 &0.04& 0.35 \\
   0.34& 0.6& 0.17
  \end{bmatrix}
  \begin{bmatrix}
   R\\
   G\\
   B
  \end{bmatrix}, 
  \end{equation}
  where $R, G,$ and $B$ are the intensity of a pixel in the red, green and blue channels, respectively. The gradient magnitude is then obtained by,
  \begin{equation}
  \label{eq_gradmag}
  ||\nabla E(i, j, \sigma)|| = \sqrt{E_{1x}^2+E_{1y}^2+E_{2x}^2+E_{2y}^2+E_{3x}^2+E_{3y}^2}, 
  \end{equation}
where $E_{kx}, E_{ky}$ are the responses of the $k$-th channel to Gaussian derivative filters in $x$ and $y$ directions given by the following impulse responses:
\begin{equation}
G_x(x, y) = \frac{-x}{2\pi \sigma^4}\exp(\frac{-(x^2+y^2)}{2\sigma^2}), 
\end{equation}
\begin{equation}
G_y(x, y) = \frac{-y}{2\pi \sigma^4}\exp(\frac{-(x^2+y^2)}{2\sigma^2}). 
\end{equation}

\begin{figure}[t]
\begin{center}
   \includegraphics[width=\linewidth]{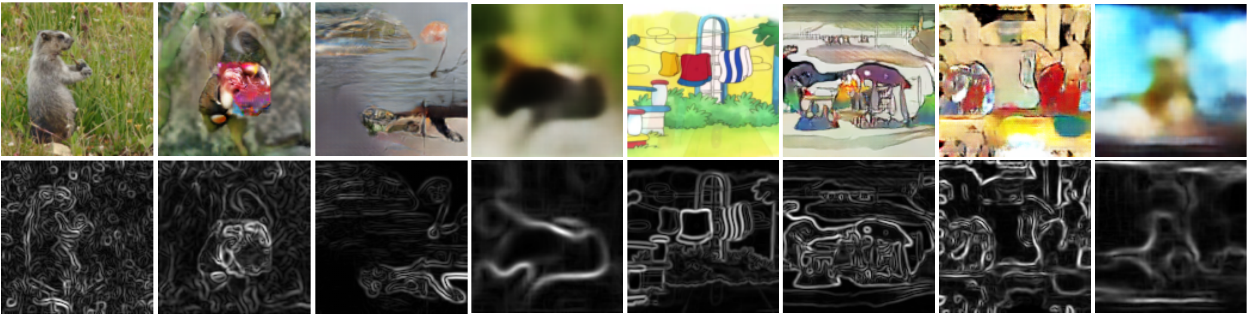}
(a)~~~~~~(b)~~~~~~(c)~~~~~~(d)~~~~~~(e)~~~~~~(f)~~~~~~(g)~~~~~~(h)
\end{center}
   \vspace{-10pt}
   \caption{Local contrast maps of (a) natural images, (b) natural images generated by DCGAN, (c) natural images generated by WGAN, (d) natural images generated by VAE, (e) cartoon images, (f) cartoon images generated by DCGAN, (g) cartoon images generated by WGAN, and (h) cartoon images generated by VAE.}
   \label{contrast_demo}
\end{figure}

The resulting gradient magnitude $||\nabla E(i, j, \sigma)||$ in eqn.\ref{eq_gradmag} is considered as local contrast of an image. Figure~\ref{contrast_demo} shows several examples of local contrast maps of training images and deep generated images. 

\subsection{Filter Responses}
\label{sec_random_filters}
It has been observed that convolving natural images with almost any zero mean linear filter results in a histogram of a similar shape with heavy tail, sharp speak and high kurtosis~\cite{zoran2013natural} (higher than kurtosis of Gaussian distribution, which is 3). That is called the non-Gaussian property of natural images.

Since it is impossible in this work to consider all these filters, we avoid inspecting responses to any specific filter. Instead, without loss of generality, we apply random zero mean filters to images as introduced in~\cite{huang1999statistics} to measure properties of images themselves. A random zero mean filter $F$ is generated by normalizing a random matrix $F_0$ with independent elements sampled uniformly from $[0, 1]$: 
\begin{equation}
F = \frac{F_0 - mean(F_0)}{||F_0 - mean(F_0)||}. 
\end{equation}

\subsection{Homogeneous Regions}
\label{sec_hom_reg}
\begin{figure}[t]
\begin{center}
   \includegraphics[width=\linewidth]{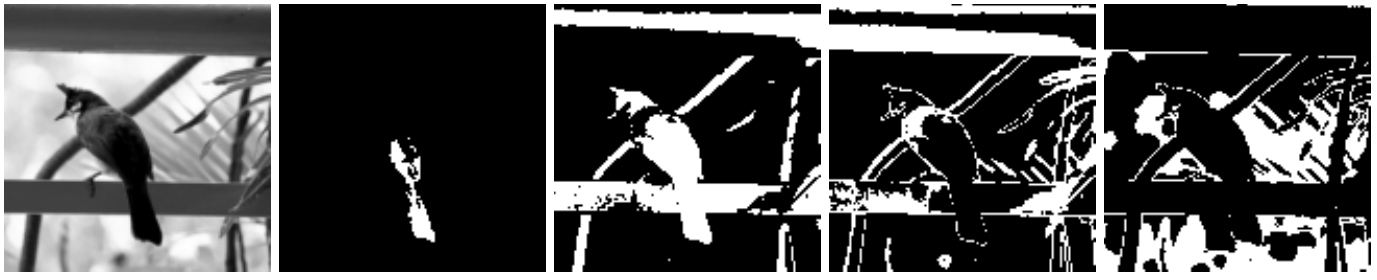}
\end{center}
   \vspace{-10pt}
   \caption{An example image and its four homogeneous regions. }
   \label{demo_components}
\end{figure}

Homogeneous regions in an image are the connected components where contrast does not exceed a certain threshold. Consider an image $I$ of size $H\times W$ and $G$ gray levels. We generate a series of thresholds $t_1, ..., t_N$, in which $t_n$ is the least integer such that more than $n\frac{HW}{N}$ pixels have a gray value less than $t_n$. Using these thresholds to segment an image results in $N$ homogeneous regions. Figure~\ref{demo_components} illustrates an example image and its homogeneous regions. 

Alvarez~\etal\cite{alvarez1999size} show that the number of homogeneous regions in natural images, denoted as $N(s)$, as a function of their size $s$, obeys the following law:
\begin{equation}
N(s) = K s^c, 
\label{alvarez_eq}
\end{equation}
where $K$ is an image dependent constant, $s$ denotes the area, and $c$ is close to -2. Suppose image $I_1$ is scaled into $I_2$, such that $I_1(a\bm{x}) = I_2(\bm{x}), a>0$. Let $N_1(s) = K_1 s^c$ denote the number of homogeneous regions of area $s$ in $I_1$. Then, for $I_2$, we have $N_2(s) = N_1(as) = K_2 s^c$, so the number of homogeneous regions in natural images is a scale-invariant statistic. 

\subsection{Power Spectrum}
We adopt the most commonly used definition of power spectrum in image statistics literature: ``the power of different frequency components". Formally, the power spectrum of an image is defined as the square of the magnitude of the image FFT. Prior studies~\cite{deriugin1956power,burton1987color,field1987relations} have shown that the mean power spectrum of natural images denoted as $S(f)$, where $f$ is frequency, is scale invariant. It has the form of:
\begin{equation}
S(f) = A f^{-\alpha}, \  \alpha \approx 2. 
\label{eq_power_spec}
\end{equation}

\section{Experiments and Results}
In this section, we report the experimental results of luminance distribution, contrast distribution, random filter response, distribution of homogeneous regions, and the mean power spectrum of the training images and deep generated images. We use Welch's t-test to test whether the statistics are significantly different between generated images and training images (ImageNet-1 and Cartoon-1 in the tables). The larger $p$-value is, the more similar the generated images to training images. Therefore, the models can be ranked according to t-test results. To make sure that our results are not specific to the choice of training images, we sampled another set of training images (ImageNet-2 and Cartoon-2 in the tables), and use t-test to measure difference between the two sets of training images. All experiments are performed using Python 2.7 and OpenCV 2.0 on a PC with Intel i7 CPU and 32GB RAM. The deep generative models used in this work are implemented in PyTorch\footnote{https://github.com/pytorch}. 

\subsection{Luminance}
\begin{figure}[htbp]
\begin{center}
\includegraphics[width=1\linewidth]{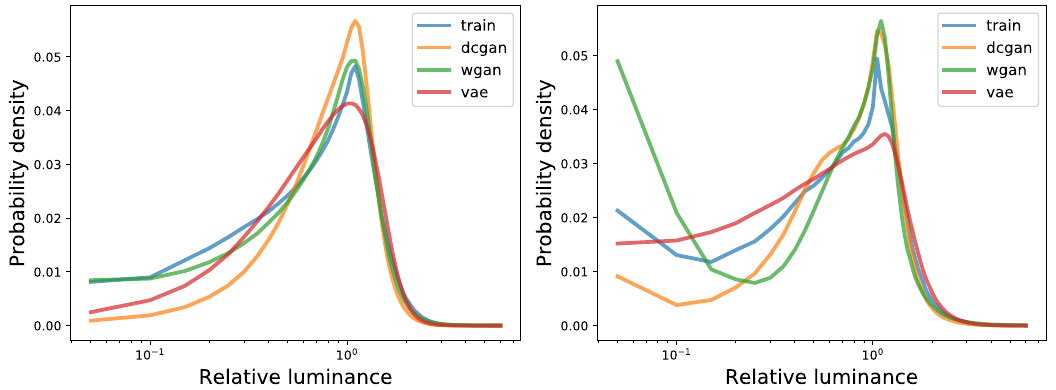}\\
(a)~~~~~~~~~~~~~~~~~~~~~~~~~~~~~~~~~~~~~~~(b)
\end{center}
   \vspace{-10pt}
   \caption{Luminance distribution. The distributions are all averaged over 12,800 images. (a) natural images and generated natural images, (b) cartoons and generated cartoon images.}
   \label{luminance}
\end{figure}

Luminance distributions of training images and deep generated images are shown in Figure~\ref{luminance}. Average skewness values are shown in Table~\ref{luminance_kurtosis}. 

Results in in Figure~\ref{luminance} show that luminance distributions of training and generated images have similar shapes, while those of cartoons are markedly different from natural images. From in Table~\ref{luminance_kurtosis}, we can see that the luminance distributions over natural images, cartoons, generated natural images and generated cartoons all have positive skewness values. However, the difference of skewness values between training and generated images is statistically significant (over both natural images and cartoons). The difference between skewness values over each image type (i.e., ImageNet-1 vs. ImageNet-2 or Cartoon-1 vs. Cartoon-2) is not significant, indicating that our our findings are general and image sets are good representatives of the natural or synthetic scenes. According to Table~\ref{luminance_kurtosis}, we rank the models in terms of luminance distribution as follows. For natural images, WGAN $>$ DCGAN $>$ VAE, and for cartoons, VAE $>$ DCGAN $>$ WGAN.
\begin{table*}[htbp]
\centering
\renewcommand{\tabcolsep}{1mm}
\caption{Skewness of the luminance distributions of the deep generated images and natural images. All values are averaged over 12,800 images. I: ImageNet, C: Cartoons.}
\label{luminance_kurtosis}
\begin{tabular}{ccccccccccc}
\toprule
-  &            I-1 & I-2 & I-DCGAN & I-WGAN &I-VAE &C-1 &C-2 &C-DCGAN &C-WGAN &C-VAE\\
\midrule
Skewness &0.1160   &0.1129   	&0.0823      &0.1435  &0.1507   &0.2987  &0.3088	&0.2316	    &0.4968  &0.2528\\
t-statistic&-     &0.3031	    &3.4506      &-2.7529 &-4.2564  &-       &-0.5139	&3.6592 	&-10.7894  &3.2696\\
p-value&    -     &0.7618		& 0.0006     &0.0059  &0.0000   &-       &0.6073	&0.0003	    &0.0000   &0.0011\\
\bottomrule
\end{tabular}
\end{table*}

\subsection{Contrast}
It has been reported that the contrast distribution in natural images follows a Weibull distribution~\cite{geusebroek2005six}. To test this on our data, first we fit a Weibull distribution (eqn.~\ref{wdis}) to the histogram of each of the generated images and training images. Then, we use KL divergence to examine if the contrast distribution in deep generated images can be well modeled by a Weibull distribution as in the case of natural images. If this is true, the fitted distributions will be close to the histogram as in training images, and thus the KL divergence will be small. 

\begin{figure*}[htbp]
\begin{center}
\includegraphics[width=0.9\linewidth]{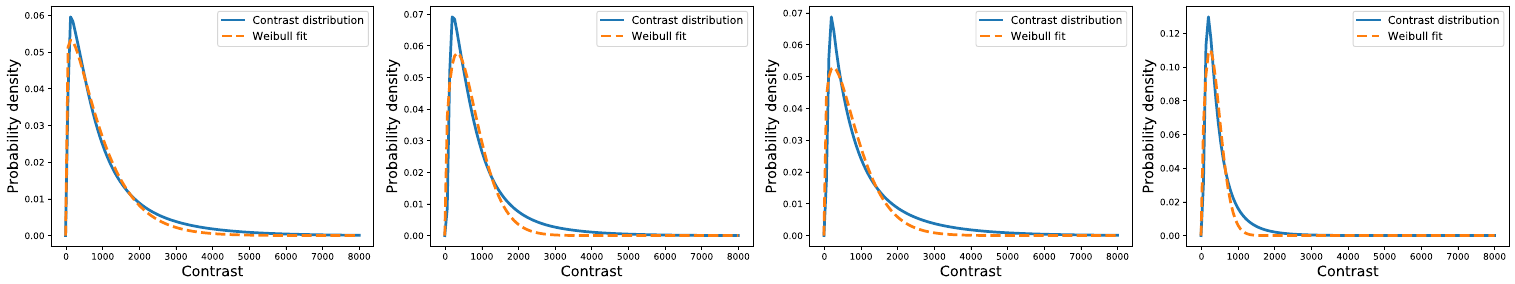}\\
\includegraphics[width=0.9\linewidth]{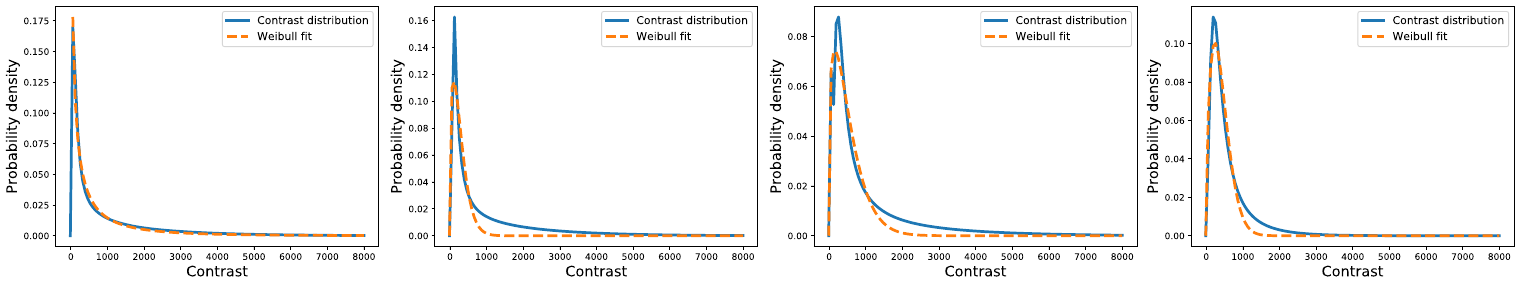}
(a)~~~~~~~~~~~~~~~~~~~~~~~~~~~~~~~(b)~~~~~~~~~~~~~~~~~~~~~~~~~~~~~~~(c)~~~~~~~~~~~~~~~~~~~~~~~~~~~~~~~(d)
\end{center}
   \vspace{-10pt}
   \caption{Contrast distributions of training images and generated images. The plots are all averaged over 12,800 images. Top: natural images, bottom: cartoons. (a) training images, (b) images generated by DCGAN, (c) images generated by WGAN, (d) images generated by VAE.}
   \label{contrast}
\end{figure*}

Figure~\ref{contrast} shows that contrast distributions of training and generated images have similar shapes, while those of cartoons are markedly different from natural images. Parameters of the fitted Weibull distribution and its KL divergence to the histogram, as well as the corresponding t-test results are shown in Table~\ref{weibull_fit}. We find that the contrast distributions in generated natural images are also Weibull distributions. However, the difference of parameters between training and generated images, in both cases of natural images and cartoons, is statistically significant. We also observe that the KL divergence between contrast distribution and its Weibull fit of natural images and generated natural images is small, while the KL divergence between contrast distribution and its Weibull fit of cartoons and generated cartoons is larger. According to Table~\ref{weibull_fit}, WGAN gets the largest $p$-value for both natural images and cartoons. DCGAN and VAE are of equally small $p$-value. Therefore, for both natural images and cartoons, WGAN $>$ DCGAN $\approx$ VAE in terms of contrast distribution.
\begin{table*}[htbp]
\centering
\renewcommand{\tabcolsep}{1mm}
\caption{Average Weibull parameters and KL divergence of training images and generated images. $\beta$ and $\gamma$ are parameters in eqn.~\ref{wdis}. All values are averaged over 12,800 images. I: ImageNet, C:cartoons.}
\label{weibull_fit}
\begin{tabular}{ccccccccccc}
\toprule
-  &            I-1 & I-2 & I-DCGAN & I-WGAN &I-VAE &C-1 &C-2 &C-DCGAN &C-WGAN &C-VAE\\
\midrule
KL divergence     &1.6845 &1.6723 &1.5005 &1.6368 &1.4986 &2.5447 &2.4914 &2.2948 &2.1382 &1.6313\\
t-statistic    & -    &1.6266 &26.6609 &6.8655 &31.9013 &-		&1.6739 &8.3537 &13.9735 &31.7103\\
p-value         & -    &0.1038  &0.0000  &0.0000  &0.0000  &-	&0.0942  &0.0000  &0.0000  &0.0000  \\
\midrule
$\gamma$       &1.1570 &1.1613 &1.2378 &1.1635 &1.1799 	&1.0115 &1.0117 &1.0211 &1.0058 &1.1305\\
t-statistic    & -    &-1.7045 &-31.1410 &-2.6475 &-10.5753	&-		&-0.2394 &-11.2113 &9.7496 &-91.8283\\
p-value         & -    &0.0883  &0.0000  &0.0081  &0.0000 	&-		&0.8108  &0.0000  &0.0000  &0.0000 \\
\midrule
$\beta$      &1251.8631 &1257.0521 &1055.3296 &1241.8656 &518.4040 	&1262.2476 &1270.1628 &1198.8230 &1247.1478 &588.8862\\
t-statistic    & -    &-1.1138 &46.9806 &2.0365 &216.7244 &-		&-1.4016 &11.6920 &2.8398 &162.8892\\
p-value         & -    &0.2654  &0.0000  &0.0417  &0.0000  &-		&0.1610  &0.0000  &0.0045  &0.0000  \\
\bottomrule
\end{tabular}
\end{table*}

\subsection{Filter Responses}
\begin{figure*}[htbp]
\begin{center}
\includegraphics[width=0.9\linewidth]{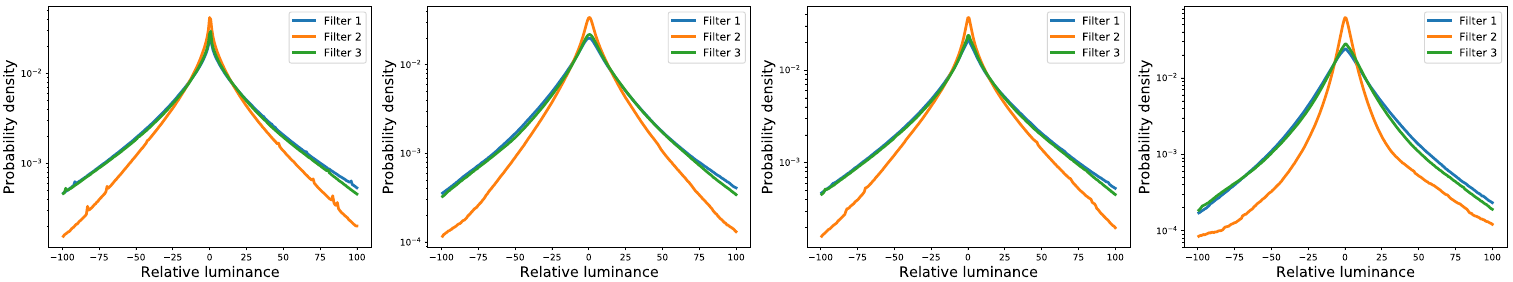}\\
\includegraphics[width=0.9\linewidth]{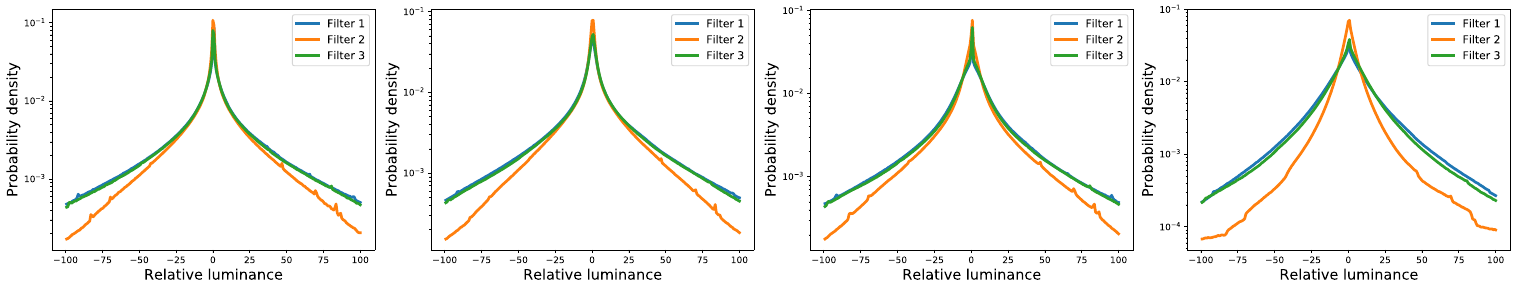}\\
(a)~~~~~~~~~~~~~~~~~~~~~~~~~~~~~~~(b)~~~~~~~~~~~~~~~~~~~~~~~~~~~~~~~(c)~~~~~~~~~~~~~~~~~~~~~~~~~~~~~~~(d)
\end{center}
   \vspace{-10pt}
   \caption{Distribution of zero mean random filters responses, averaged over 12,800 images. Top: natural images, bottom: cartoons. (a) training images, (b) images generated by DCGAN, (c) images generated by WGAN, and (d) images generated by VAE.}
   \label{zmr_filter}
\end{figure*}

We generate three $8\times8$ zero mean random filters as in~\cite{huang1999statistics}, and apply them to ImageNet training images, VAE generated images, DCGAN generated images and WGAN generated images. Averaging the response histograms over training images, VAE images, DCGAN images and WGAN images gives the distributions shown in Figure~\ref{zmr_filter} (in order). The distributions of responses to different random filters have similar shapes with a sharp peak and a heavy tail, which is in agreement with Huang~\etal's results~\cite{huang1999statistics}. Average kurtosis of the filter response distributions over the training images and deep generated images are shown in Table~\ref{zmr_filter_kurtosis}. 

Figure~\ref{zmr_filter} shows that generated images have similar filter response distributions to training images, while those of cartoons looks different from natural images. Table~\ref{zmr_filter_kurtosis} shows that the average responses kurtosis of generated natural images and real natural images are all greater than 3 of Gaussian distribution. As a result, we draw the conclusion that the generated natural images also have similar non-Gaussianity as in natural images. However, there is a statistically significant difference of the filter response kurtosis between deep generated images and training images, in both cases of natural images and cartoons (except ImageNet-WGAN and Cartoon-DCGAN). For natural images, WGAN gets the largest $p$-value of filter 1, 2 and 3. DCGAN and VAE are of similar $p$-value. Therefore for natural images, WGAN $>$ DCGAN $\approx$ VAE. For cartoons, WGAN gets the largest $p$-value of filter 3, and DCGAN gets the largest $p$-value of filter 1, and 2. Therefore, for cartoons, DCGAN $>$ WGAN $>$ VAE.
\begin{table*}[htbp]
\centering
\renewcommand{\tabcolsep}{1mm}
\caption{Kurtosis of the distributions of responses to three zero mean random filters. I: ImageNet, C: Cartoons.}
\label{zmr_filter_kurtosis}
\begin{tabular}{ccccccccccc}
\toprule
-  &            I-1 & I-2 & I-DCGAN & I-WGAN &I-VAE &C-1 &C-2 &C-DCGAN &C-WGAN &C-VAE\\
\midrule
filter 1 &       5.9315     &5.9117 		&6.7088    			&5.8742    &11.3657  &7.9789	&7.9711	&7.9786	&8.2128	&11.3245 \\
t-statistic    & -          &1.4255         &-16.4376      	     &1.2623   &-101.3270  &-	&0.0599	&0.0029	&-2.1637	&-31.1486\\
p-value         & -         &0.6705         & 0.0000      	     &0.2068   &0.0000  &-	&0.9522	&0.9977	&0.0305	 &0.0000 \\
\midrule
filter 2 &        5.7729    &5.6780   		&7.1181    				&5.9575 &15.7773    &7.3999	&7.4632	&7.4325	&7.1798	 & 12.0194 \\
t-statistic    & -          &1.0885         &-29.0011      	     & -3.9459  &-175.0824   &-			&-0.6474	&-0.3326	&2.7206 &-56.4231 	\\
p-value         & -         &0.0368         & 0.0000      	     & 0.0001   &0.0000  &-			&0.5174	&0.7395	&0.0065	&0.0000 \\
\midrule
filter 3 &         5.7901    &5.7412  		 &6.1794     &5.7747   & 11.9817   &5.1254	&5.1916	&5.3312	&5.1954	& 8.4338\\
t-statistic    & -           &1.0504        &-8.4558     & 0.3381  &-110.7246   &-	    &-0.4870	&-1.7311	&-0.6991	&-35.5865 \\
p-value         & -          &0.2936        & 0.0000     & 0.7353  &0.0000   &-	    &0.6263	&0.0834	  &0.4845 &0.0000 	\\
\bottomrule
\end{tabular}
\end{table*}

\subsection{Homogeneous Regions}
\begin{figure*}[htbp]
\begin{center}
\includegraphics[width=0.9\linewidth]{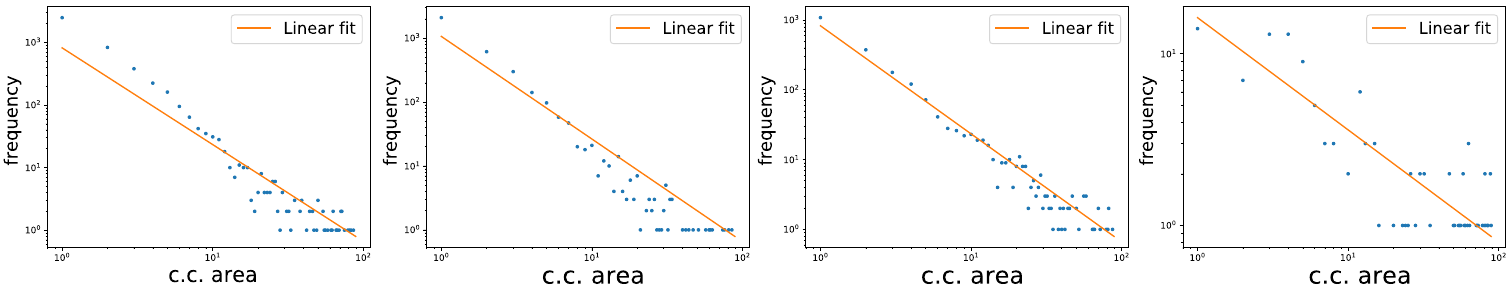}\\
\includegraphics[width=0.9\linewidth]{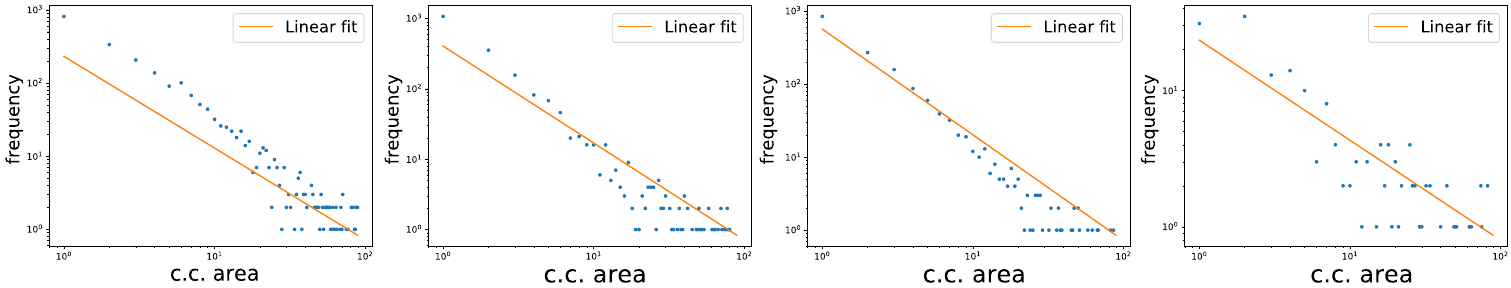}\\
(a)~~~~~~~~~~~~~~~~~~~~~~~~~~~~~~~(b)~~~~~~~~~~~~~~~~~~~~~~~~~~~~~~~(c)~~~~~~~~~~~~~~~~~~~~~~~~~~~~~~~(d)
\end{center}
   \vspace{-10pt}
   \caption{The number of homogeneous regions of area $s$ in training images and generated images (both axes are in log units). The plots are all averaged over 12,800 images. Top: natural images, bottom: cartoons. (a) training images, (b) images generated by DCGAN, (c) images generated by WGAN, and (d) images generated by VAE.}
   \label{cc_area}
\end{figure*}

We compute distribution of homogeneous regions as stated in Section~\ref{sec_hom_reg}. The number of homogeneous regions ($N$ in Section~\ref{sec_hom_reg}) is set to 16. Figure~\ref{cc_area} shows the distribution of the number of the homogeneous regions of area $s$ in the training images and deep generated images. We use eqn.~\ref{alvarez_eq} to fit the distribution of homogeneous regions of each image. Table~\ref{cc_area_param} shows average parameters $K$ and $c$ in eqn.~\ref{alvarez_eq} evaluated through maximum likelihood (only regions of area $s<90$ pixels are considered in the evaluation). 

Over real natural images and generated natural images, the relationship between the number and the area of regions is linear in log-log plots, thus supporting the scale invariance property observed by Alvarez~\etal\cite{alvarez1999size}. This is also reflected from the small fitting residual to eqn.~\ref{alvarez_eq} shown in the first column of Table~\ref{cc_area_param}. We also find this property holds over cartoons and generated cartoons. However, the differences between the parameters of deep generated images and training images, in both cases of natural images and cartoons, are statistically significant (except ImageNet-WGAN). For natural images, WGAN has the largest $p$-value. DCGAN and VAE are of equally small $p$-value. Therefore, we rank the models for natural images as follow: WGAN $>$ DCGAN $\approx$ VAE. For cartoons, three models have similar $p$-value, therefore, WGAN $\approx$ DCGAN $\approx$ VAE.
\begin{table*}[htbp]
\centering
\renewcommand{\tabcolsep}{1mm}
\caption{Parameters $K$ and $c$ in eqn.~\ref{alvarez_eq} computed using maximum likelihood. I: ImageNet, C: Cartoons.}
\label{cc_area_param}
\begin{tabular}{ccccccccccc}
\toprule
-  &            I-1 & I-2 & I-DCGAN & I-WGAN &I-VAE &C-1 &C-2 &C-DCGAN &C-WGAN &C-VAE\\
\midrule
$c$ &         -1.5459      	&-1.5521		&   -1.6051    		&  -1.5484  & -0.6535    &-1.2541	&-1.2527	&-1.3791	&-1.4476  &-0.7333	\\
t-statistic    & -          &2.2611        & 23.1626     	     & 0.9645   &-387.7215 	&-			&-0.4930	&52.8988	&82.6709  &-220.1396	\\
p-value         & -         &0.0238         & 0.0000      	     & 0.3348   & 0.0000	&-			&0.6220		&0.0000	&0.0000	      &0.0000 \\

\midrule
$K$ &          2.9135      &2.9234			&   3.0268     			& 2.9176   & 1.2102    &2.3657	&2.3622	&2.9130	&2.7534	 &1.3687  \\
t-statistic    & -         &-2.0333          & -25.2192      	     &-0.8973   & 411.3727	&-		&0.6277	&-51.3570	&-83.1429& 212.5199\\
p-value         & -        &0.0420          & 0.0000      	     	& 0.3696   & 0.0000	&-		&0.5302	&0.0000	&0.0000      &	0.0000\\

\midrule
residual&   3.8795       &3.8844			&   3.8299     			&3.8962   & 3.9824   &4.2407	&4.2253	&4.3959	&4.1129	&3.8593 \\
t-statistic    & -       &-0.4125            & 4.4927      	     & -1.5561   &-8.9440	&-			&1.2462	&-13.2940	&11.6040	&34.1295\\
p-value         & -      &0.6800            & 0.0000      	    	 &0.1197    &0.0000	&-			&0.2127	&0.0000	&0.0000	&0.0000\\
\bottomrule
\end{tabular}
\end{table*}

\subsection{Power Spectrum}
\begin{figure*}[htbp]
\begin{center}
\includegraphics[width=0.9\linewidth]{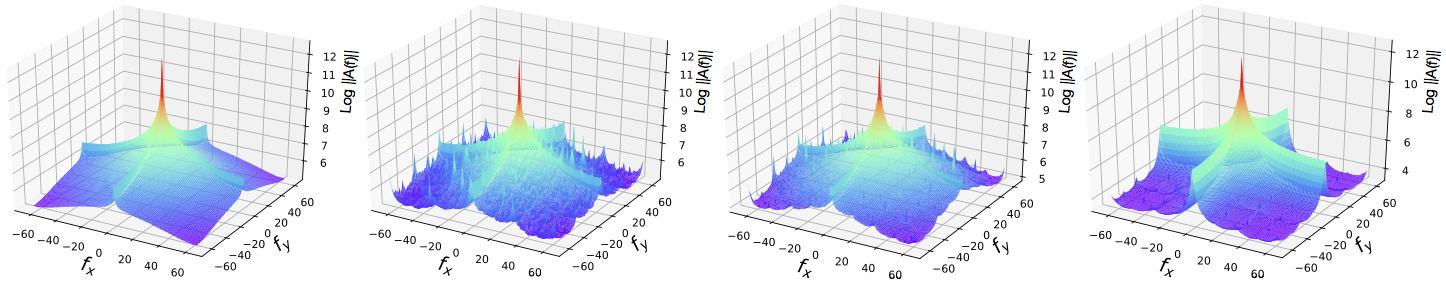}\\
\includegraphics[width=0.9\linewidth]{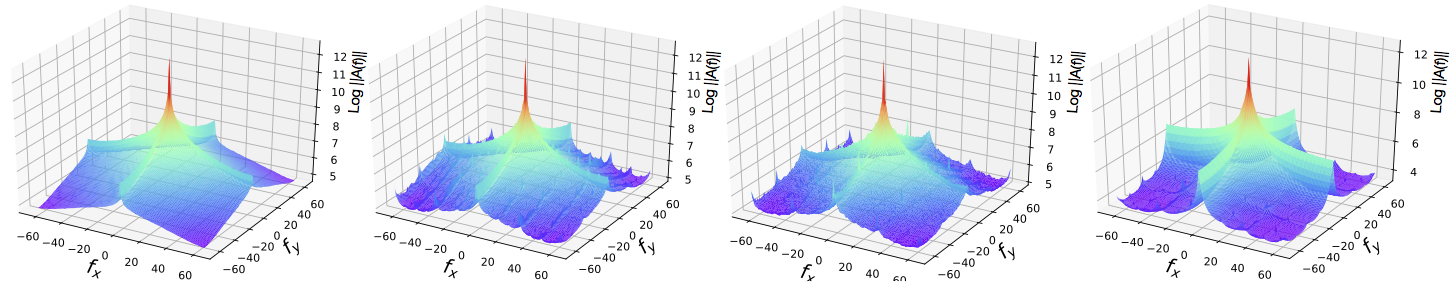}\\
(a)~~~~~~~~~~~~~~~~~~~~~~~~~~~~~~~(b)~~~~~~~~~~~~~~~~~~~~~~~~~~~~~~~(c)~~~~~~~~~~~~~~~~~~~~~~~~~~~~~~~(d)
\end{center}
   \vspace{-10pt}
   \caption{Mean power spectrum averaged over 12800 images. Magnitude is in logarithm unit. Top: natural images, bottom: cartoons. (a) training images, (b) images generated by DCGAN, (c) images generated by WGAN, and (d) images generated by VAE.}
   \label{power_spec}
\end{figure*}

\begin{figure*}[htbp]
\begin{center}
\includegraphics[width=0.9\linewidth]{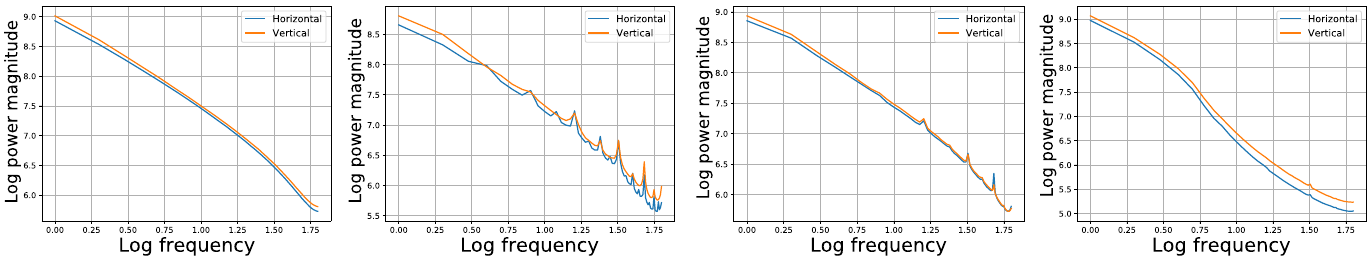}\\
\includegraphics[width=0.9\linewidth]{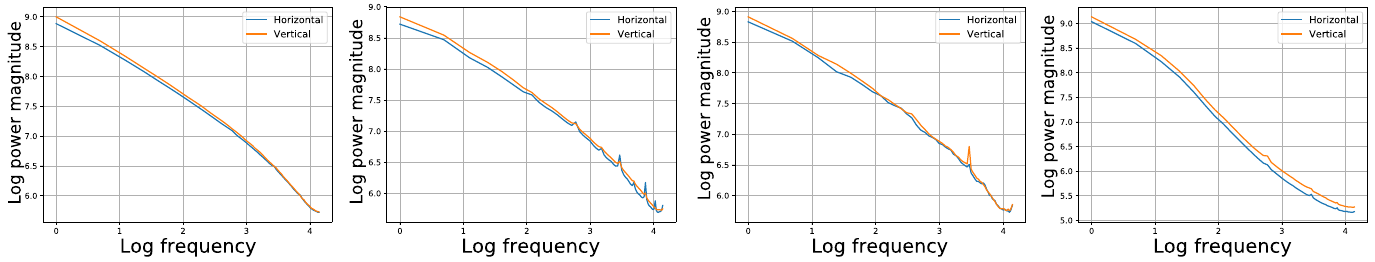}\\
(a)~~~~~~~~~~~~~~~~~~~~~~~~~~~~~~~(b)~~~~~~~~~~~~~~~~~~~~~~~~~~~~~~~(c)~~~~~~~~~~~~~~~~~~~~~~~~~~~~~~~(d)
\end{center}
   \vspace{-10pt}
   \caption{Mean power spectrum averaged over 12800 images in horizontal and vertical directions. Both magnitude and frequency are in logarithm unit. Top: natural images, bottom: cartoons. (a) training images, (b) images generated by DCGAN, (c) images generated by WGAN, and (d) images generated by VAE.}
   \label{log_power_spec2d}
\end{figure*}

Figures~\ref{power_spec}(a), and~\ref{log_power_spec2d}(a) show the mean power spectrum of training images. We use eqn.~\ref{eq_power_spec} to fit a power spectrum to each image. The evaluated parameters of eqn.~\ref{eq_power_spec} and the corresponding fitting residual, averaged over all images, are shown in Table~\ref{power_param}. The results of t-test show that differences between the parameters of deep generated images and training images, in both cases of natural images and cartoons, are statistically significant. For natural images, WGAN has the largest $p$-value. DCGAN and VAE are of similar $p$-value, therefore, for natural images, WGAN $>$ DCGAN $\approx$ VAE. For cartoons, WGAN has the largest $p$-value of $\alpha_h$ and $A_v$. VAE has the largest $p$-value of residual-v. DCGAN has the largest $p$-value of residual-h. Therefore, for cartoons, WGAN $>$ DCGAN $\approx$ VAE. 

Scale invariant mean power spectrum of natural images with the form of eqn.~\ref{eq_power_spec} is the earliest and the most robust discovery in natural image statistics. Our experiments on training images confirm this discovery and aligns with the prior results in~\cite{deriugin1956power, cohen1975image, burton1987color, field1987relations}. This can be seen from the linear relationship between log frequency and log power magnitude shown in Figure~\ref{log_power_spec2d}(a), and the small fitting residual to eqn.\ref{eq_power_spec} in the first column of Table~\ref{power_param}. We also observe a similar pattern over cartoons. 

However, unlike training images, the generated images seem to have a spiky mean power spectrum. See Figure~\ref{power_spec}(b)(c) and Figure~\ref{log_power_spec2d}(b)(c). It can be seen from the figures that there are several local maxima of energy in certain frequency points. Without frequency axis being taken logarithm, it can be read from Figure~\ref{power_spec}(b)(c) that the position of each spike is the integer multiple of $\frac{4}{128}$ cycle/pixel. 
\begin{table*}[htbp]
\centering
\renewcommand{\tabcolsep}{1mm}
\caption{Fitted parameters of eqn.~\ref{eq_power_spec} from the mean power spectra (averaged over 12800 images in horizontal and vertical directions). $A_h$, $\alpha_h$, residual-h: the parameters and fitting residual of horizontally averaged power spectrum; $A_v$, $\alpha_v$, residual-v: parameters and fitting residual of vertically averaged power spectrum. I: ImageNet, C: Cartoon.}
\label{power_param}
\begin{tabular}{ccccccccccc}
\toprule
-  &            I-1 & I-2 & I-DCGAN & I-WGAN &I-VAE &C-1 &C-2 &C-DCGAN &C-WGAN &C-VAE\\
\midrule
$A_h$&9.2383 &9.2418 &8.9781 &9.2113 &8.8342   &9.1690 &9.1649 &9.0792 &9.2039 &8.9417	\\
t-statistic&-	&-0.7504 &57.9637 &6.0118 &105.7172  & -         &0.6150 &14.1713 &-6.1758 &43.9667 	\\
p-value &-	&0.4530  &0.0000  &0.0000  &0.0000       & -   &0.5386  &0.0000  &0.0000  &0.0000 		\\

\midrule
$\alpha_h$&-1.9733 &-1.9733 &-1.8882 &-1.9542 &-2.3313	 &-1.9813 &-1.9768 &-1.9461 &-1.9869 &-2.3234	\\
t-statistic&-		&0.0262 &-26.6477 &-5.7127 &111.1663  & -  &-1.2210 &-9.8188 &1.6845 &102.4380	\\
p-value &-		 &0.9791  &0.0000  &0.0000  &0.0000     & -   &0.2221  &0.0000  &0.0921  &0.0000 	\\

\midrule
residual-h&1.5341 &1.5523 &2.5745 &1.4119 &2.1860  &2.0091 &2.0202 &2.0621 &1.7371 &1.8183 \\
t-statistic&-	&-1.2755 &-61.2857 &10.6709 &-52.3226   & - &-0.5625 &-2.9619 &18.1469 &12.7869	\\
p-value  &-		&0.2021  &0.0000  &0.0000  &0.0000   & -&0.5738  &0.0031  &0.0000  &0.0000 \\

\midrule
$A_v$ &9.2909 &9.2884 &9.0066 &9.2886 &8.9713  &9.2804 &9.2794 &9.1861 &9.2792 &9.1315	\\
t-statistic &-	&0.5775 &67.4247 &0.5441 &86.9562   & -  &0.1637 &16.2791 &0.2273 &29.4495 	\\
p-value  &-	&0.5636  &0.0000  &0.5864  &0.0000    & - &0.8700  &0.0000  &0.8202  &0.0000  	\\

\midrule
$\alpha_v$&-1.9592 &-1.9562 &-1.8165 &-1.9748 &-2.3382  &-2.0327 &-2.0295 &-1.9845 &-2.0197 &-2.4213		\\
t-statistic &-	&-0.8855 &-45.6106 &5.0362 &105.7561  & -  &-0.9126 &-14.5242 &-4.2188 &96.7694  	\\
p-value  &-		&0.3759  &0.0000  &0.0000  &0.0000     & -  &0.3615  &0.0000  &0.0000  &0.0000 	\\

\midrule
residual-v &1.4563 &1.4656 &1.7456 &1.4804 &2.1362 &1.9864 &1.9836 &1.8448 &1.8995 &1.9976	\\
t-statistic &-	&-0.6300 &-21.2443 &-2.0089 &-49.1039  & -  &0.1285 &8.0197 &5.4635 &-0.6603 	\\
p-value    &-	&0.5287  &0.0000  &0.0446  &0.0000   & -  &0.8978  &0.0000  &0.0000  &0.5091 	\\
\bottomrule
\end{tabular}
\end{table*}

\begin{figure}[htbp]
\begin{center}
\includegraphics[width=0.9\linewidth]{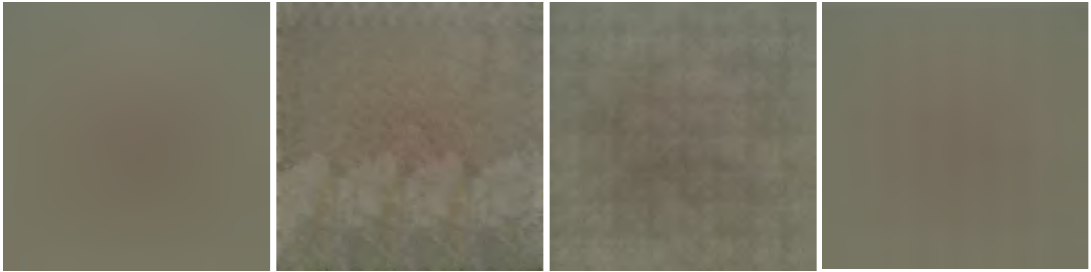}\\
(a)~~~~~~~~~~~~~~~(b)~~~~~~~~~~~~~~~(c)~~~~~~~~~~~~~~~(d)
\end{center}
   \vspace{-10pt}
   \caption{Average images of training and generated images (each of them are averaged over 12,800 images). The average generated images show periodic patterns. (a) training images, (b) images generated by DCGAN, (c) images generated by WGAN, and (d) images generated by VAE.}
   \label{ave_img}
\end{figure}

It is surprising to see that unlike natural images, deep generated images do not meet the well-established scale invariance property of the mean power spectrum. These models, however, well reproduce other statistical properties such as Weibull contract distribution and non-Gaussianity. Specifically, mean power magnitude $||A(f)||$ of natural images falls smoothly with frequency $f$ of the form $||A(f)|| \propto 1/f^\alpha$, but mean power spectra of the deep generated images turns out to have local energy maxima at the integer multiples of frequency of 4/128 (i.e., 4/128, 8/128, etc). In spatial domain, this indicates that there are some periodic patterns with period of $32, 16, ..., 4, 2$ pixels superimposed on the generated images. Averaging the deep generated images gives an intuitive visualization of these periodic patterns. Please see Figure~\ref{ave_img}. 

\section{Summary and Conclusion}
We explore statistics of images generated by state-of-the-art deep generative models (VAE, DCGAN and WGAN) and cartoon images with respect to the natural image statistics.
Our analyses on training natural images corroborates existing findings of scale invariance, non-Gaussianity, and Weibull contrast distribution on natural image statistics. We also find non-Gaussianity and Weibull contrast distribution for generated images with VAE, DCGAN and WGAN. These statistics, however, are still significantly different. Unlike natural images, neither of the generated images has scale invariant mean power spectrum magnitude, which indicates extra structures in the generated images. 

Inspecting how well the statistics of the generated images match natural scenes, can a) reveal the degree to which deep learning models capture the essence of the natural scenes, b) provide a new dimension to evaluate models, and c) suggest possible directions to improve image generation models. Correspondingly, two possible future works include: 
\begin{enumerate}
\item Building a new metric for evaluating deep image generative models based on image statistics, and 
\item Designing deep image generative models that better capture statistics of the natural scenes (e.g., through designing new loss functions).
\end{enumerate}

To encourage future explorations in this area and assess the quality of images by other image generation models, we share our cartoon dataset and code for computing the statistics of images at: \href{https://github.com/zengxianyu/generate}{https://github.com/zengxianyu/generate.}



%



\section*{Acknowledgment}
The authors would like to thank several teachers at DUT for their helpful comments on an earlier version of this work. We wish to thank Mr. Yingming Wang for providing some cartoon videos, and Dr. Lijun Wang for his help. 


\ifCLASSOPTIONcaptionsoff
  \newpage
\fi



%
\bibliographystyle{ieee}
\bibliography{reference}

\end{document}